\documentclass[10pt,twocolumn,letterpaper]{article}

\usepackage{cvpr}              

\usepackage[ruled]{algorithm2e}
\usepackage{listings}
\usepackage{graphicx}
\usepackage{amsmath, bm}
\usepackage{amssymb}
\usepackage{booktabs}
\usepackage{multirow}
\usepackage[super]{nth}
\usepackage{color}
\usepackage[export]{adjustbox}
\usepackage{array}
\usepackage[accsupp]{axessibility}
\newcolumntype{P}[1]{>{\centering\arraybackslash}p{#1}}

\definecolor{caribbeangreen}{rgb}{0.0, 0.8, 0.6}

\usepackage[pagebackref,breaklinks,colorlinks]{hyperref}
\usepackage[dvipsnames]{xcolor}

\usepackage[capitalize]{cleveref}
\crefname{section}{Sec.}{Secs.}
\Crefname{section}{Section}{Sections}
\Crefname{table}{Table}{Tables}
\crefname{table}{Tab.}{Tabs.}

\def\cvprPaperID{10480} 
\def\confName{CVPR}
\def\confYear{2023}

\begin{document}

\title{Generalization Matters: Loss Minima Flattening via Parameter Hybridization for Efficient Online Knowledge Distillation}
\makeatletter
\newcommand{\myfnsymbol}[1]{%
  \expandafter\@myfnsymbol\csname c@#1\endcsname
}
\newcommand{\@myfnsymbol}[1]{%
  \ifcase #1
  \or \TextOrMath{\textdagger}{\dagger}
  \fi
}

\author{
    Tianli Zhang\textsuperscript{$1$},
    Mengqi Xue\textsuperscript{$2$},
    Jiangtao Zhang\textsuperscript{$1$},
    Haofei Zhang\textsuperscript{$1$},
    Yu Wang\textsuperscript{$1$},
    Lechao Cheng\textsuperscript{$3$}, \\
    Jie Song\textsuperscript{$1,\dagger$},
    and Mingli Song\textsuperscript{$1$} \\
    \textsuperscript{$1$}Zhejiang University,
    \textsuperscript{$2$}Hangzhou City University,
    \textsuperscript{$3$}Zhejiang Lab\\
     {\tt\small \{zhangtianli,zhjgtao,haofeizhang,yu.wang,sjie,brooksong\}@zju.edu.cn},\\
     {\tt\small mqxue@zucc.edu.cn}, 
     {\tt\small chenglc@zhejianglab.com}\\
}
\renewcommand{\thefootnote}{\myfnsymbol{footnote}}
\maketitle
\footnotetext[1]{Corresponding author}%
\setcounter{footnote}{0}
\renewcommand{\thefootnote}{\arabic{footnote}}
\maketitle

\begin{abstract}
    Most existing online knowledge distillation~(OKD) techniques typically require sophisticated modules to produce diverse knowledge for improving students' generalization ability.
    In this paper, we strive to fully utilize multi-model settings instead of well-designed modules to achieve a distillation effect with excellent generalization performance.
    Generally, model generalization can be reflected in the flatness of the loss landscape.
    Since averaging parameters of multiple models can find flatter minima, we are inspired to extend the process to the sampled convex combinations of multi-student models in OKD. 
    Specifically, by linearly weighting students' parameters in each training batch, we construct a Hybrid-Weight Model~(HWM) to represent the parameters surrounding involved students.
    The supervision loss of HWM can estimate the landscape's curvature of the whole region around students to measure the generalization explicitly.
    Hence we integrate HWM's loss into students' training and propose a novel OKD framework via parameter hybridization~(OKDPH) to promote flatter minima and obtain robust solutions.
    Considering the redundancy of parameters could lead to the collapse of HWM, we further introduce a fusion operation to keep the high similarity of students.
    Compared to the state-of-the-art~(SOTA) OKD methods and SOTA methods of seeking flat minima, our OKDPH achieves higher performance with fewer parameters, benefiting OKD with lightweight and robust characteristics. 
    Our code is publicly available at \href{https://github.com/tianlizhang/OKDPH}{https://github.com/tianlizhang/OKDPH}.
\end{abstract}

\section{Introduction} \label{sec:intro} 
Deep learning achieves breakthrough progress in a variety of tasks by constructing a large capacity network pre-trained on massive data~\cite{dong2021survey}.
In order to apply high-parameterized models in the real-world scene with limited resources, the knowledge distillation~(KD) technique \cite{hinton2015distilling} aims to obtain a compact and effective student model guided by a large-scale teacher model for model compression. 
Based on the developments of KD, Zhang \etal \cite{zhang2018deep} propose the concept of online knowledge distillation~(OKD) to view all networks as students and achieve mutual learning from scratch through peer teaching, liberating the distillation process from the dependency on pre-trained teachers.
Existing OKD methods mainly encourage students to acquire diverse and rich knowledge, including aggregating predictions~\cite{guo2020online, su2021attention, wang2021adaptable}, combining features~\cite{kim2021feature, li2022embedded, phuong2019distillation}, working with peers~\cite{wu2021peer, zhu2022online}, learning from group leaders~\cite{chen2020online}, and receiving guidance from online teachers~\cite{wu2021peer}.

Nevertheless, these strategies focus on designing sophisticated architectures to exploit heterogeneous knowledge to enhance students' generalization, but they lack explicit constraints on generalization.
The concept of generalization to deep models is the ability to fit correctly on previously unseen data~\cite{neyshabur2017exploring}, which can be reflected by the flatness of the loss landscape~\cite{keskar2016large, hochreiter1997flat}.
Flatness is the landscape's local curvature, which is costly to direct calculate by the Hessian.
Considering the setting of multiple students in OKD, we utilize the theory of multi-model fusion in parameter space~\cite{frankle2020linear} to estimate the local curvature by the linear combination of students' parameters (we call it a \textit{hybrid-weight model}, which is expressed as HWM).
More specifically, HWM is a stochastic convex combination of parameters of multiple students on different data augmentations, which can sample multiple local points on the landscape.
Intuitively, HWM's loss reflects the upper and lower bounds of the local region's loss and estimates the curvature of the landscape.
Minimizing HWM's loss flattens the region and forms a landscape with smaller curvature.

\begin{figure}[t]
  \centering
  \includegraphics[width=0.8\linewidth]{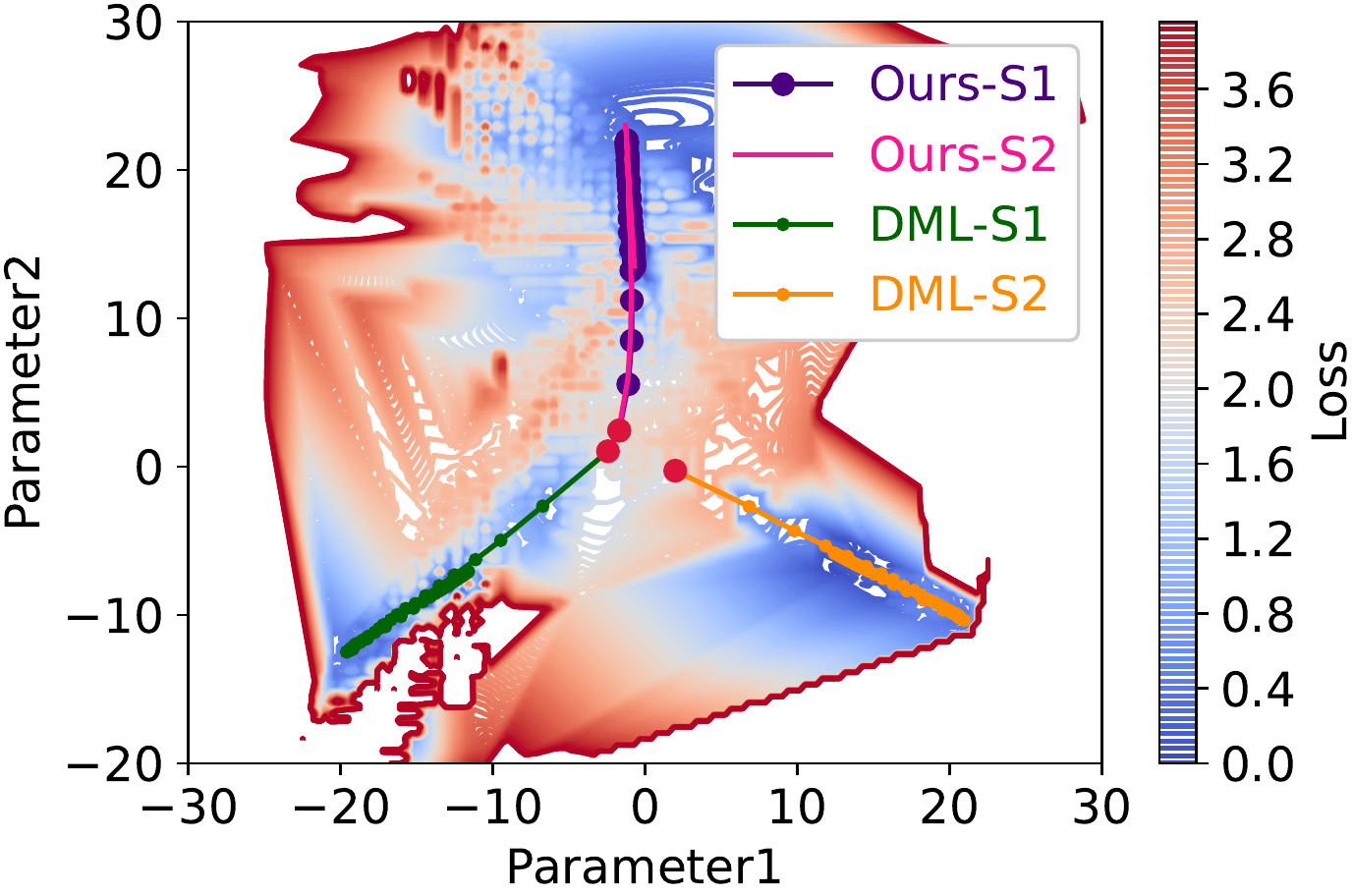}
  \caption{The loss landscape visualization of four students~(\textcolor{Plum}{Ours-S1} and \textcolor{VioletRed}{Ours-S2} are obtained by our method, and DML obtains \textcolor{ForestGreen}{DML-S1} and \textcolor{BurntOrange}{DML-S2}), which are ResNet32 \cite{he2016deep} trained by the same settings on CIFAR-10 \cite{krizhevsky2009learning}.
  Four students start from the initial point~(Red points in the center) and converge to three basins along different trajectories.
  The x-axis and y-axis represent the values of model parameters that PCA~\cite{mackiewicz1993principal} obtains.
  }
  \label{fig:loss_contour}
\end{figure}

Based on the above observation, we propose a concise and effective OKD framework, termed \textit{online knowledge distillation with parameter hybridization}~(OKDPH), to promote flatter loss minima and achieve higher generalization.
We devise a novel loss function for students' training that incorporates the standard Cross-Entropy~(CE) loss and Kullback-Leibler~(KL) divergence loss, but also a supervised learning loss from HWM.
Specifically, HWM is constructed in each batch by linearly weighting multiple students' parameters.
The classification error of HWM explicitly measures the flatness of the region around students on the loss landscape, reflecting their stability and generalization.
The proposed loss equals imposing stronger constraints on the landscape, guiding students to converge in a more stable direction.
For intuitive understanding, we visualize the loss landscape of students obtained by different methods in \cref{fig:loss_contour}.
Our students converge to one broader and flatter basin~(thus superior generalization performance), while the students obtained by DML \cite{zhang2018deep} converge to different sharp basins, degrading the robustness and performance.

Unfortunately, directly hybridizing students' parameters can easily lead to the collapse of HWM due to the high nonlinearity of deep neural networks~\cite{neyshabur2020being, singh2020model}.
Therefore, we restrict the differences between students through intermittent fusion operations to ensure the high similarity of multi-model parameters and achieve effective construction of HWM.
Concretely, at regular intervals, we hybridize the parameter of HWM with each student and, conversely, assign the hybrid parameter to the corresponding student.
This process shortens the distance between students, shown as very close loss trajectories of our students in \cref{fig:loss_contour}.
However, it will not reduce diversity because students receive different types of data augmentation, and they can easily become diverse during training.
Our method pulls students in the same direction, plays the role of strong regularization, and obtains one lightweight parameter that performs well in various scenarios.
The solution derived from our method is expected to integrate the dark knowledge from multiple models while maintaining a compact architecture and can be competent for resource-constrained applications.

To sum up, our contributions are organized as follows:
\begin{itemize}
    \item Inspired by the theory of multi-model fusion, we innovatively extend traditional weight averaging to an on-the-fly stochastic convex combination of students' parameters, called a hybrid-weight model~(HWM).
    The supervision loss of HWM can estimate the curvature of the loss landscape around students and explicitly measure the generalization.
    \item We propose a brand-new extensible and powerful OKD framework via parameter hybridization~(OKDPH) for loss minima flattening, which flexibly adapts to various network architectures without modifying peers' structures and extra modules.
    It is the first OKD work that manipulates parameters.
    \item Extensive experiments on various backbones demonstrate that our OKDPH can considerably improve the students' generalization and exceed the state-of-the-art~(SOTA) OKD methods and SOTA approaches of seeking flat minima.
    Further loss landscape visualization and stability analysis verify that our solution locates in the region having uniformly low loss and is more robust to perturbations and limited data.
\end{itemize}

\begin{figure*}[t]
  \centering
  \includegraphics[width=1.0\linewidth]{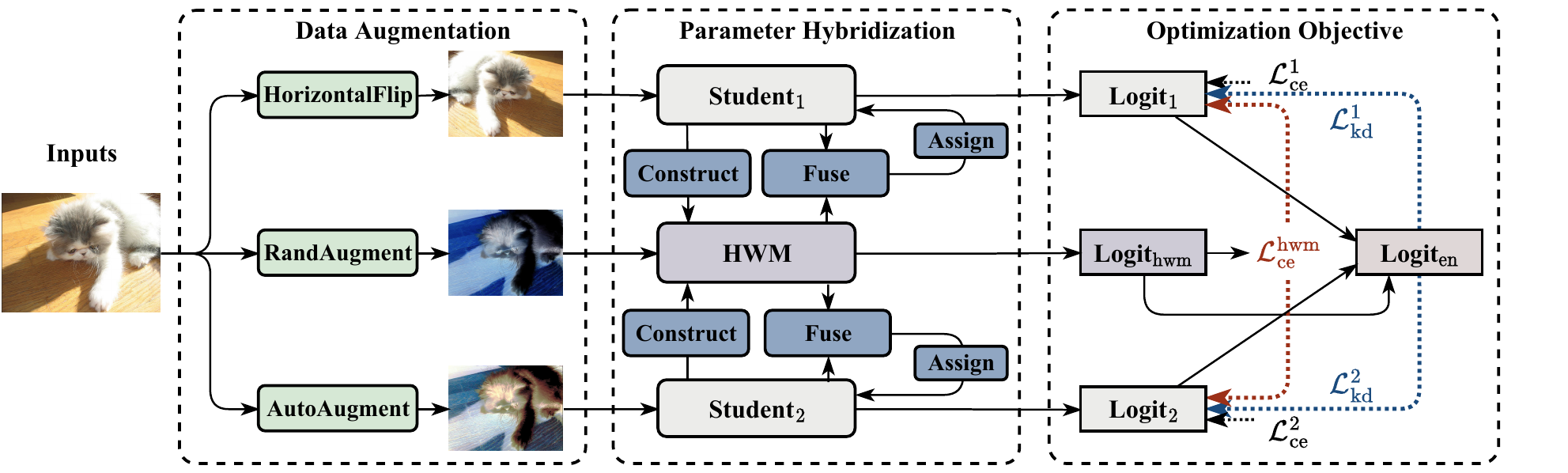}
  \caption{
    Framework of our OKDPH. Two students construct HWM by sampled convex parameter combinations in each training batch, and HWM's parameters are regularly fused with students.
    Two students and HWM's logits are obtained by feeding three types of data augmentations and are averaged to $\text{Logit}_{en}$.
    Each student's training loss consists of the classification loss $\mathcal{L}_{ce},\mathcal{L}_{ce}^{hwm}$ and the KD loss $\mathcal{L}_{kd}$.
    }
  \label{fig:framework}
\end{figure*}

\section{Related Work}
\label{sec:formatting}

\paragraph{Online Knowledge Distillation.} 
Zhang \etal \cite{zhang2018deep} propose deep mutual learning~(DML) that enables students to share knowledge from each other's predictions to achieve distillation without teachers.
KDCL \cite{guo2020online}, which is improved on DML, integrates the output of multiple students under different data augmentations as soft labels to guide students to optimize.
In contrast to DML and KDCL, which promote collaboration between several students, ONE \cite{zhu2018knowledge} employs a gating component to achieve distillation under the framework of one head and auxiliary branches.
OKDDip \cite{chen2020online} incorporates the attention mechanism into the multi-branch structure and guides students to learn from their auxiliary peers and the group leader.
To constrain hidden representation between sub-networks, FFL \cite{kim2021feature} constructs a feature fusion module to improve the distillation effect, while PCL \cite{wu2021peer} builds a temporal mean model to act as an online teacher to produce more stable predictions.
Ding \etal~\cite{ding2021knowledge} design a knowledge refinery~(KR) pipeline with decoupled labels to eliminate extra cumbersome teachers.

\paragraph{Generalization.} 
Hinton \etal \cite{hinton1993keeping} apply the minimum description length principle to study the relationship between the generalization and the sharpness of minima.
Hochreiter \etal \cite{hochreiter1997flat} and Keskar \etal \cite{keskar2016large} propose that the flatness of the loss landscape basin nearby the solution is an indicator to measure the model generalization ability.
According to Dziugaite \etal \cite{dziugaite2017computing}, the average empirical error is small if the model lies in a flat region of the parameter space.
Langford \etal \cite{langford2001not} construct the distribution of the model and improve its generalization by sensitivity analysis, that is, adding Gaussian noise with variance each time evaluating the data.
Neyshabur \etal \cite{neyshabur2017exploring} associate sharpness with PAC-Bayes and believe network scales, such as norm and margin, also affect the generalization ability.
In addition, Mobahi \etal \cite{mobahi2020self} demonstrate how distillation can improve the generalization ability of networks through regularization and sparsity in Hilbert space.

\paragraph{Parameter Fusion.} 
Frankle \etal \cite{frankle2020linear} define a phenomenon called linear mode connectivity~(LMC) as the parameters of two networks a path can connect with low loss.
They hold that if two models can be linearly connected without barriers, they are inclined to be in the same basin and show more stability to noise.
Model Soup \cite{wortsman2022model} uses this principle to average parameters of multiple pre-trained models and achieves remarkable performance improvement on ImageNet.
Neyshabur \etal \cite{neyshabur2020being} believe that two networks, even with the same random initialization, can be observed LMC barriers. 
Singh \etal \cite{singh2020model} propose that since there is no one-to-one correspondence between well-trained network layers, achieving model fusion by direct average parameters is challenging.
Tarvainen \etal~\cite{tarvainen2017mean} construct a mean teacher with more accurate labels by continuous students' exponential moving average~(EMA), resulting in better test accuracy.
SWA~\cite{izmailov2018averaging} achieves a wide and generalizable solution by weighted averaging the local minimum located in the border of areas with lower loss.
Considering the relationship between the loss landscape's geometry and generalization, SAM~\cite{foret2020sharpness} seeks parameters in neighborhoods with uniformly low loss.

\section{Method}
\subsection{Vanilla Online Knowledge Distillation}

Generally, vanilla OKD replaces the commonly used pre-trained teachers with peer student models. The training loss consists of the Cross-Entropy (CE) loss and the Knowledge Distillation (KD) loss.
Let $\mathcal{D}=\{x_i, y_i\}_{i=1}^{N}$ be a training set containing $N$ images and $C$ categories of labels. The $m$-th student ($m \in \{1, \cdots, M\}$) obtains its output logits $\mathbf{z}^m \in \mathcal{R^C}$ by feeding an augmentation of $x_i$.
The classification loss is calculated by Cross-Entropy:
\begin{equation}
    {\mathcal{L}_{ce}}\left({\mathbf{z}^m}, y_i\right) = - \log \frac {\exp \left({\mathbf{z}^m_{y_i}}\right)} {\sum_{c=1}^{C} \exp \left({\mathbf{z}_c^{m}}\right)},\label{eq:ce}
\end{equation}
where $y_i \in \{1, \cdots, C\}$ is the ground-truth label of the image $x_i$.
The KD loss requires measuring the alignment of output distribution between models, which is usually achieved by KL divergence with temperature:
\begin{equation}
    {\mathcal{L}}_{kd} = \tau^2 \mathcal{D}_{KL} \left(p^m,p^j\right)=\tau^2 \sum_{c=1}^{C}{p^j_c \log\frac{p^j_c}{p^m_c}},
\end{equation}
where $p^m, p^j \in \mathcal{R^C}$ are the soft logits produced by a pair of students $m$ and $j$.
The soft logits are calculated by:
\begin{equation}
    p^m=\sigma({{\mathbf{z}}^m}/{\tau})=\frac{\exp({{\mathbf{z}}^m}/{\tau})}{\sum_c {\exp({{\mathbf{z}}^m_c}/{\tau})} },
\end{equation}
where $\sigma$ is the softmax function, and $\tau$ is the temperature.

\subsection{Parameter Hybridization}
It is known that a trained model is a point in weight space, and averaging multiple points leads to finding flatter minima in the loss landscape.
Considering the multi-model setting in OKD, we extend the average process to constructing a hybrid-weight model~(HWM) using a sampled convex parameter combination of peer students.
Formally, we create an HWM in each batch during training as follows:
\begin{equation}
        {\theta}_{hwm}^t \mathbf{=} \sum_{m=1}^{M}{{r}_m^t {{\theta}_m^t}},\
        \mathbf{r}^t=[r_1^t, ..., r_M^t] \sim \text{Dir}\left( \mathbf{\alpha}\right)
        \label{eq:hwm-construct}
\end{equation}
where $t$ represents the $t$-th training bath. ${\theta}^{t}_{hwm}$ and ${\theta}^{t}_m$ are HWM's and the $m$-th student's parameters at the $t$-th training batch, $M$ is the number of students, and $\mathbf{r}^t$ is the weight vector that subjects to $\sum_{m=1}^M r_m^t=1$.

$\text{Dir}\left( \mathbf{\alpha}\right)$ is the Dirichlet distribution parameterized by $\mathbf{\alpha} \in \mathcal{R}^M$, which is commonly adopted as a prior distribution for multivariate sampling~\cite{lda}.
Dynamic sampling $\mathbf{r}^t$ subject to $\text{Dir}\left( \mathbf{\alpha}\right)$ can achieve the following effects: 
first, HWM can fully explore the parameter points of the region around students by each batch sampling;
second, the concentration vector $\mathbf{\alpha}$ can easily adjust the sampling probability of different points in this region.
Consistent with the relevant works of averaging multiple models~\cite{izmailov2018averaging,wortsman2022model,frankle2020linear}, we pay more attention to the parameter center of multiple models.
We fix $\mathbf{\alpha}=\mathbf{1} \in \mathcal{R}^M$ to gradually increase the sampling probability from the borderline models to the center of gravity.
As shown in~\cref{fig:parameter_update}, the color of the triangles represents the probability distribution of the 3-dimensional Dirichlet distribution $\text{Dir}\left( [1,1,1]\right)$.
The darker the color, the higher the sampling probability.

Previous studies show that over-parameterized students, even under the supervision of KD loss, are prone to stay away from each other during training, resulting in the collapse of their parameter hybridization~\cite{neyshabur2020being,singh2020model}. 
So we require an additional operation to constrain the similarity between students to construct HWM effectively.
Inspired by some works about parameter fusion~\cite{tarvainen2017mean,jing2021amalgamating}, we fuse HWM and students in a particular proportion at regular intervals during training:
\begin{equation}
    \text{If mod}\left(t,\Delta\right)=0:\
    {\theta}_{m}^{t} \mathbf{=} {\gamma} {\theta}_{hwm}^{t} + (1-{\gamma}) {\theta}_{m}^{t},
    \label{eq:student-fusion}
\end{equation}
where $\Delta$ and $\gamma$ are two hyper-parameters to represent the fusion interval and ratio, respectively.
$\Delta$ can be set at the epoch or batch level, for instance, every epoch or every five batches. 
$\gamma$ is generally taken as $0.5$ or $1$, which means the average of HWM and each student or directly replacing each student with HWM, respectively.
When $\gamma=1$, the role of the fusion is to achieve an instant parameter ensemble of students through HWM in the training process.

In short, constructing HWM and the fusion between HWM and students make up our parameter hybridization strategy.
The former samples the points of the region around the students in the parameter space, and the latter controls the range of this region to prevent students from diverging.

\subsection{Optimization Objective and Procedure}
\cref{fig:framework} shows the framework of our OKDPH in the case of two student models.
Like some multi-model studies~\cite{guo2020online, wortsman2022model}, the student models and their HWM receive different data augmentation to produce informative and diverse predictions.
Naturally, we synthesize knowledge in various scenarios by averaging the output logits of each model:
\begin{equation}
    \mathbf{z}^{en} \mathbf{=} \frac{1}{M+1} \left( \sum_{m=1}^{M}{\mathbf{z}^m} + \mathbf{z}^{hwm} \right),
    \label{eq:logits-ensemble}
\end{equation}
where $\mathbf{z}^m$ and $\mathbf{z}^{hwm}$ are the $m$-th students' and HWM's logits, respectively.
The holistic knowledge of the ensemble logits $ \mathbf{z}^{en}$ is further distilled into each student model by:
\begin{equation}
     \mathcal{L}_{kd}({\mathbf{z}}^m, {\mathbf{z}}^{en}) =  \tau^2 \mathcal{D}_{KL}\left( \sigma \left(\frac{{\mathbf{z}}^m}{\tau} \right), \sigma\left(\frac{{\mathbf{z}}^{en}}{\tau} \right) \right) \label{eq:kd},
\end{equation}
where $\mathcal{D}_{KL}$ is the KL divergence between the $m$-th student's logits $\mathbf{z}^m$ and the ensemble logits $\mathbf{z}^{en}$.

As mentioned in the previous section, the sampled HWM represents surrounding parameters in the space spanned by peer students. 
The HWM's classification loss ${\mathcal{L}}_{ce}^{hwm}$ for multiple consecutive training batches reflects upper and lower bounds on the loss in the region around students.
Hence we integrate ${\mathcal{L}}_{ce}^{hwm}$ into each student's training to minimize the curvature of the region and flatten the loss landscape, enhancing students' generalization ability:
\begin{equation}
    \mathcal{L}^m = { \omega {\mathcal{L}}_{ce}^m + (1-\omega) {\mathcal{L}}_{ce}^{hwm} + \beta {\mathcal{L}}_{kd}({\mathbf{z}}^m, {\mathbf{z}}^{en}) },
    \label{eq:full-loss}
\end{equation}
where ${\mathcal{L}}^m$ and ${\mathcal{L}}_{ce}^m$ are the $m$-th student's total loss and CE loss, respectively.
The loss term $\omega$ can adjust the proportion of the current student's loss and its surrounding model's loss in the total classification loss.
$\beta$ represents the ratio of KD loss relative to CE loss, usually the same value as $\omega$.

\begin{figure}[t]
  \centering
  \includegraphics[width=0.85\linewidth]{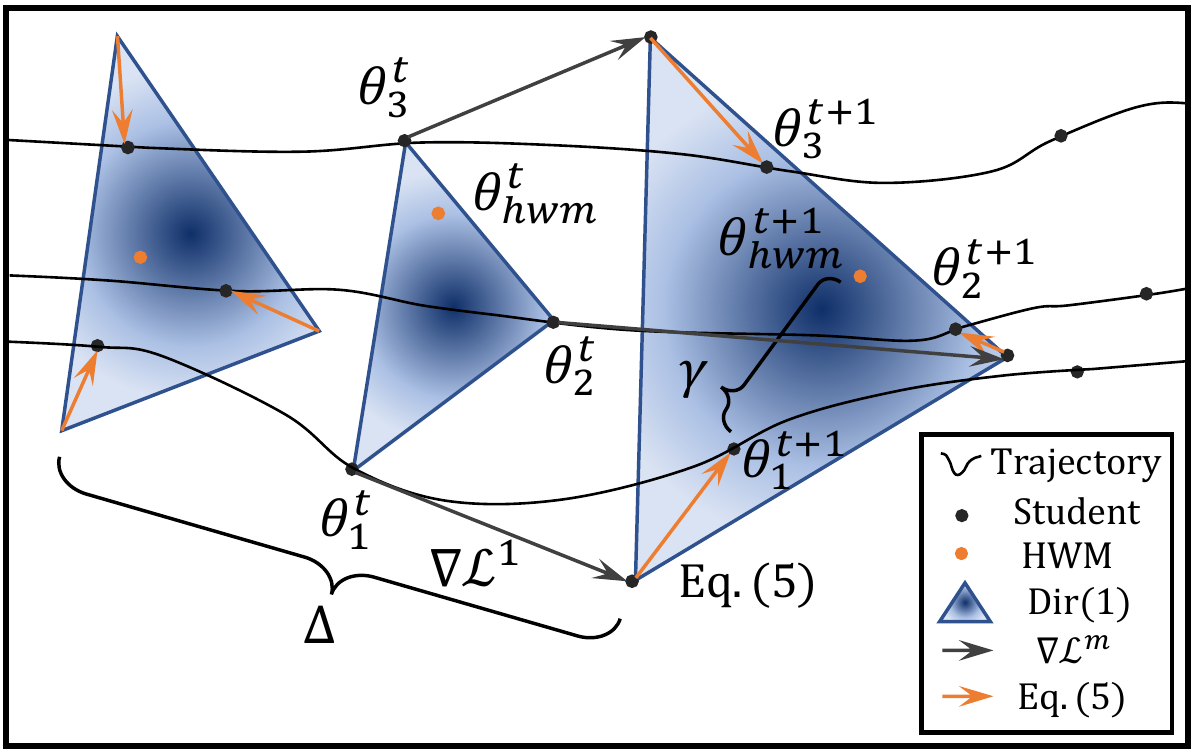}

  \caption{
  Schematic of the OKDPH parameter update.
  }
  \label{fig:parameter_update}
\end{figure}

Our OKDPH procedure is summarized in~\cref{alg:algorithm1}, and~\cref{fig:parameter_update} shows the parameter update trajectories of three student models.
Specifically, three students in the $t$-th training batch, \ie, $\theta_1^t,\theta_2^t,\theta_3^t$, are optimized to $\theta_1^{t+1},\theta_2^{t+1}$ and $\theta_3^{t+1}$ by the gradient of~\cref{eq:full-loss}.
Then, their HWM $\theta_{hwm}^{t+1}$ is sampled on the triangle, formed by three student points. 
In the next training batch, the gradient of HWM's classification loss is passed to each student and guides the students to optimize to flatter loss minima in the back-propagation process.
Notably, three students will gradually move away from each other in the parameter space, shown as a gradually larger triangle in~\cref{fig:parameter_update}.
So every time the interval $\Delta$ is reached, we reduce the distance of three students by the fusion with HWM, shown as an orange arrow.

\begin{algorithm}[t]
	\caption{OKD with Parameter Hybridization.}
	\label{alg:algorithm1}

    \KwIn{Training data and labels: $\mathcal{D}=\{x_i, y_i\}_{i=1}^N$;\\
    Number of training batches and students: $T$, $M$;\\
    Students and augmentations: $\{\theta_m^1\}_{m=1}^M$, $\{H_m\}_{m=1}^{M+1}$;\\
    Fusion interval and ratio: $\Delta, \gamma$; Loss terms: $\omega, \beta$.}
	\KwOut{The best model parameter $\theta_{best}$.}  
	
	\BlankLine

	Initialize HWM $\theta_{hwm}^1$ by~\cref{eq:hwm-construct};\\

    \ForEach{$t \in \{1, \cdots, T\}$}{

		Sample a batch of images and labels ${X},{Y} \sim \mathcal{D}$;
		
        \ForEach{$m \in \{1,\cdots,M\}$}{
		    Predict logits output $\mathbf{z}^m \mathbf{=} \theta_m^t(H_m(X))$;
		    
		}
        Predict logits output $\mathbf{z}^{hwm} \mathbf{=} \theta_{hwm}^t(H_{M+1}(X))$;
  
		Compute ensemble predictions $\mathbf{z}^{en}$ by~\cref{eq:logits-ensemble};
		
		\ForEach{$m \in \{1,\cdots,M\}$}{
            Optimize ${\theta_m^{t}}$ to ${\theta_m^{t+1}}$ by gradient of ~\cref{eq:full-loss};

		}
		
		Construct HWM $\theta_{hwm}^{t+1}$ by~\cref{eq:hwm-construct};

        \ForEach{$m \in \{1,\cdots,M\}$}{
		    Update $\theta_m^{t+1}$ by~\cref{eq:student-fusion};
            
		}
	}
	\Return Best model ${\theta}_{best} \in \{\theta_1^t, ..., \theta_M^t, \theta_{hwm}^t\}_{t=1}^T$.

\end{algorithm}

\section{Experiment}
We first describe the datasets and experimental settings in~\cref{settings} and compare the proposed OKDPH with SOTA methods in~\cref{results}.
Then~\cref{generalization:measure} measures the generalization by the loss landscape visualization and stability analysis.
Also, we conduct the parameter sensitivity and ablation study in~\cref{parameter_sensitivity} and~\cref{ablation:analyse}.

\subsection{Experimental Settings} \label{settings}
\noindent\textbf{Datasets.}
\textbf{CIFAR-10} \cite{krizhevsky2009learning} and \textbf{CIFAR-100} \cite{krizhevsky2009learning}, as two commonly-used small-scale datasets for OKD, are adopted to verify the effectiveness of OKDPH. Moreover, we also introduce a large-scale benchmark, \ie, \textbf{ImageNet} \cite{russakovsky2015imagenet}, for validating our method with complex real-world images.

\begin{table*}[htbp]
\setlength{\tabcolsep}{2.0pt}
\begin{center}
\small
\begin{tabular}{c|ccccc|ccccc}
\toprule
\textbf{Dataset} & \multicolumn{5}{c|}{\footnotesize \textbf{CIFAR-10}} & \multicolumn{5}{c}{\footnotesize \textbf{CIFAR-100}} \\
\midrule
\footnotesize \textbf{Method} & \footnotesize \textbf{ResNet32} & \footnotesize \textbf{ResNet110} &
    \footnotesize \textbf{VGG16} & \footnotesize \textbf{DenseNet40-12} & \footnotesize \textbf{WRN20-8}
    & \footnotesize \textbf{ResNet32} & \footnotesize \textbf{ResNet110} & \footnotesize \textbf{VGG16} & 
    \footnotesize \textbf{DenseNet40-12} & \footnotesize \textbf{WRN20-8} \\
\midrule
\footnotesize \textbf{EMA} & \footnotesize $93.71_{\pm0.07}$ & $94.78_{\pm0.22}$ & \footnotesize $94.07_{\pm0.10}$  & \footnotesize $93.25_{\pm0.16}$  & \footnotesize $94.68_{\pm0.16}$ & \footnotesize $71.29_{\pm0.31}$ & \footnotesize $75.26_{\pm0.67}$ & \footnotesize $72.64_{\pm0.13}$ & \footnotesize $70.82_{\pm0.27}$ & \footnotesize $76.58_{\pm0.17}$ \\
\footnotesize \textbf{SWA} & \footnotesize $94.00_{\pm0.09}$ & \footnotesize $94.79_{\pm0.15}$ & \footnotesize $94.56_{\pm0.04}$ & \footnotesize $93.18_{\pm0.14}$ & \footnotesize $94.65_{\pm0.08}$ & \footnotesize $72.06_{\pm0.32}$ & \footnotesize $75.26_{\pm0.49}$ & \footnotesize $74.69_{\pm0.18}$ & \footnotesize $71.10_{\pm0.36}$ & \footnotesize $76.31_{\pm0.27}$ \\
\footnotesize \textbf{SAM} & \footnotesize $94.41_{\pm0.17}$ & \footnotesize $94.91_{\pm0.20}$  & \footnotesize $95.13_{\pm0.07}$ & \footnotesize  $93.17_{\pm0.30}$ & \footnotesize $95.13_{\pm0.15}$ & \footnotesize $72.54_{\pm0.17}$ & $75.50_{\pm0.43}$ & \footnotesize $74.55_{\pm0.15}$ & \footnotesize $70.74_{\pm0.36}$ & \footnotesize $75.86_{\pm0.19}$ \\
\midrule
\footnotesize \textbf{DML}    & \footnotesize $94.27_{\pm0.08}$ & \footnotesize $95.13_{\pm0.14}$ & \footnotesize $94.28_{\pm0.13}$ & \footnotesize $93.66_{\pm0.05}$ & \footnotesize $95.04_{\pm0.14}$ & \footnotesize $72.82_{\pm0.17}$ & \footnotesize $75.91_{\pm0.23}$ & \footnotesize $73.56_{\pm0.05}$ & \footnotesize $71.57_{\pm0.21}$ & \footnotesize $77.29_{\pm0.13}$ \\
\footnotesize \textbf{ONE}    & \footnotesize $94.31_{\pm0.07}$ & \footnotesize $95.27_{\pm0.13}$ & \footnotesize $93.83_{\pm0.07}$ & \footnotesize $92.95_{\pm0.06}$ & \footnotesize $94.79_{\pm0.03}$ & \footnotesize $74.02_{\pm0.29}$ & \footnotesize $78.37_{\pm0.30}$ & \footnotesize $72.59_{\pm0.15}$ & \footnotesize $70.32_{\pm0.23}$ & \footnotesize $77.98_{\pm0.33}$ \\
\footnotesize \textbf{KDCL}   & \footnotesize $93.91_{\pm0.08}$ & \footnotesize $95.11_{\pm0.16}$ & \footnotesize $94.24_{\pm0.08}$ & \footnotesize $93.87_{\pm0.08}$ & \footnotesize $95.27_{\pm0.16}$ & \footnotesize $71.83_{\pm0.34}$ & \footnotesize $78.28_{\pm0.32}$  & \footnotesize $73.98_{\pm0.22}$  & \footnotesize $71.35_{\pm0.42}$  & \footnotesize $77.97_{\pm0.30}$  \\
\footnotesize \textbf{OKDDip} & \footnotesize $94.19_{\pm0.05}$ & \footnotesize $95.16_{\pm0.16}$  & \footnotesize $93.72_{\pm0.48}$ & \footnotesize $93.03_{\pm0.27}$  & \footnotesize $94.61_{\pm0.08}$  & \footnotesize $71.71_{\pm0.18}$  & \footnotesize $77.60_{\pm0.20}$  & \footnotesize $72.71_{\pm0.19}$  & \footnotesize $70.30_{\pm0.18}$  & \footnotesize $77.75_{\pm0.15}$  \\
\footnotesize \textbf{FFL}    & \footnotesize $94.32_{\pm0.07}$ & \footnotesize $95.33_{\pm0.18}$  & \footnotesize $93.92_{\pm0.09}$ & \footnotesize $92.93_{\pm0.15}$  & \footnotesize $92.16_{\pm0.12}$  & \footnotesize $73.39_{\pm0.32}$  & \footnotesize $77.61_{\pm0.25}$  & \footnotesize $72.95_{\pm0.21}$  & \footnotesize $70.78_{\pm0.29}$  & \footnotesize $69.53_{\pm0.14}$  \\
\footnotesize \textbf{PCL}    & \footnotesize $94.20_{\pm0.12}$ & \footnotesize $94.85_{\pm0.33}$  & \footnotesize $94.22_{\pm0.04}$ & \footnotesize $92.25_{\pm0.32}$  & \footnotesize $92.57_{\pm0.24}$  & \footnotesize $72.86_{\pm0.31}$  & \footnotesize $78.33_{\pm0.25}$  & \footnotesize $73.54_{\pm0.33}$  & \footnotesize $69.95_{\pm0.24}$  & \footnotesize $77.88_{\pm0.24}$  \\
\footnotesize \textbf{KR}  & \footnotesize $93.37_{\pm0.19}$ & \footnotesize $95.19_{\pm0.26}$ & \footnotesize $93.77_{\pm0.14}$ & \footnotesize $92.64_{\pm0.13}$ & \footnotesize $94.93_{\pm0.18}$ & \footnotesize $70.12_{\pm0.07}$ & \footnotesize $75.10_{\pm0.61}$ & \footnotesize $72.21_{\pm0.16}$ & \footnotesize $69.31_{\pm0.43}$ & \footnotesize $74.66_{\pm0.16}$ \\
\midrule
\footnotesize \textbf{OKDPH}  & \footnotesize $\pmb{95.01_{\pm0.09}}$ & \footnotesize $\pmb{96.28_{\pm0.09}}$ & \footnotesize $\pmb{95.32_{\pm0.05}}$ & \footnotesize $\pmb{94.34_{\pm0.13}}$ & \footnotesize $\pmb{95.95_{\pm0.09}}$ & \footnotesize $\pmb{74.10_{\pm0.22}}$ & \footnotesize $\pmb{79.68_{\pm0.29}}$ & \footnotesize $\pmb{75.56_{\pm0.12}}$ & \footnotesize $\pmb{72.30_{\pm0.26}}$ & \footnotesize $\pmb{78.88_{\pm0.08}}$ \\
\bottomrule
\end{tabular}
\caption{Top 1 accuracy~(\%) and standard deviation comparison of SOTA methods on CIFAR-10 and CIFAR-100.}
\label{CIFAR}
\end{center}
\vspace{-.5em}
\vspace{-.5em}
\end{table*}

\begin{table}[t]
\setlength{\tabcolsep}{2pt}
\begin{center}
\small
\begin{tabular}{l|cccccccc}
\toprule
\footnotesize \textbf{Backbone} 
& \footnotesize \textbf{KD} & \footnotesize \textbf{DML} & \footnotesize \textbf{ONE} & \footnotesize \textbf{KDCL}  & \footnotesize \textbf{OKDDip} & \footnotesize \textbf{FFL} & \footnotesize \textbf{PCL}  & \footnotesize \textbf{Ours} \\
\midrule
\footnotesize \textbf{ResNet18} 
& \footnotesize 69.51 & \footnotesize 69.82 & \footnotesize 70.18 & \footnotesize 69.60  & \footnotesize 69.93 & \footnotesize 68.85 & \footnotesize 70.22  & \footnotesize \textbf{70.66} \\
\bottomrule
\end{tabular}
\caption{Accuracy~(\%) of several OKD methods on ImageNet.}
\label{ImageNet}
\end{center}
\end{table}

\noindent\textbf{Backbones and Training Details.}
We use the framework of PyTorch~\cite{paszke2019pytorch} to implement our experiment in the setting of two student models and provide results with more students in the supplementary materials.
For CIFAR-10 and CIFAR-100 datasets, we evaluate OKDPH on students with various backbones, including ResNet32 \cite{he2016deep}, ResNet110 \cite{he2016deep}, VGG16 \cite{szegedy2015going}, DenseNet40-12 \cite{huang2017densely}, and WRN20-8 \cite{zagoruyko2016wide}.
Each model receives multiple combinations of data augmentations, including random crop and normalization. 
Except for the above data augmentations, the two students and HWM also accept random horizontal flipping, Cutout \cite{devries2017improved}, and Random Augment \cite{cubuk2020randaugment}, respectively.
The SGD optimizer~\cite{sutskever2013importance} is adopted with a learning rate of $0.1$ and a weight decay of $5e^{-4}$.
The number of epochs and the batch size are set to 300 and 128, respectively.
For the settings of ImageNet, we employ the standard ResNet18 \cite{he2016deep} as the backbone and train 100 epochs with a learning rate of $0.1$.

\noindent\textbf{Baselines.}
The compared baselines include the mainstream methods of seeking flat minima and SOTA OKD methods, which aim at validating the superiority of our OKDPH in both two research fields.
The former kind of methods includes exponential moving average~(EMA~\cite{tarvainen2017mean}), stochastic weight averaging~(SWA~\cite{izmailov2018averaging}), and sharpness aware minimization~(SAM~\cite{foret2020sharpness}).
As for OKD methods, DML \cite{zhang2018deep} and KDCL \cite{guo2020online} encourage students' cooperation to promote mutual learning, while  ONE \cite{zhu2018knowledge}, OKDDip \cite{chen2020online}, FFL \cite{kim2021feature}, PCL \cite{wu2021peer}, and KR~\cite{ding2021knowledge} design additional modules or labels to produce and employ meaningful knowledge.

\subsection{Results and Analysis} 
\label{results}
As shown in \cref{CIFAR} and \cref{ImageNet}, we compare the top-1 accuracy of the proposed OKDPH with several SOTA methods on three datasets
Like other OKD works and for fair comparisons, we report the best accuracy of one single model on the testing set and show the standard deviation by averaging five consecutive runs with a fixed random seed.

First, our method has the most outstanding performance on CIFAR-10, which is $0.64\%$, $1.00\%$, $0.20\%$, $0.50\%$, and $0.71\%$ higher than the SOTA method on the backbones of ResNet32, ResNet110, VGG16, DenseNet40-12, and WRN20-8 respectively.
Strikingly, the proposed method breaks through the $95\%$ accuracy rate on ResNet32 in the OKD field for the first time. 
Secondly, on CIFAR-100, it can be observed that our OKDPH consistently beats all SOTA methods, which outperforms the sub-optimal method by $1.67\%$ and $1.16\%$ on the ResNet110 and VGG16.
It is no exaggeration to say that OKDPH raises the upper limit of OKD's performance on two CIFAR datasets.
Also, our OKDPH performs significantly better than SOTA methods of seeking flat minima because our method utilizes extra multifarious knowledge brought by multiple models and the loss of distillation.
Last but not least, we conduct experiments on ImageNet ILSVRC 2012 \cite{russakovsky2015imagenet} to further verify the usefulness of OKDPH in scenarios involving sizable datasets. 
\cref{ImageNet} illustrates that OKDPH achieves the best top-1 accuracy compared with other OKD methods.

From the perspective of knowledge distillation, we believe that part of the reason for OKDPH's success is the broad scope of knowledge interaction.
The carrier of knowledge transfer in OKD is usually the logits~\cite{zhang2018deep,guo2020online,xue2021kdexplainer} and layers’ representation~\cite{kim2021feature,yang2022deep,yang2022factorizing}.
We realize the former by~\cref{eq:kd} and extend the latter to the parameter level, which improves the output of all layers, not just the top-level layer.
On the one hand, we conjecture that the parameter hybridization strategy can automatically integrate various dark knowledge~\cite{hinton2015distilling,shen2019customizing} encoded in multiple students, unlike other OKD baselines that rely on well-designed modules to generate effective knowledge~\cite{zhu2018knowledge,wu2021peer,song2021tree,ding2021knowledge}.
On the other hand, considering that students receiving different data augmentations can hybridize parameters without harming the accuracy, our OKDPH can be viewed as a regularization operation that regulates each layer's functionality and latent representations~\cite{mobahi2020self,zhang2020self}, leading to one lightweight parameter that performs well in various scenarios.
In summary, our method achieves higher performance with fewer parameters than the SOTA baselines in both fields, with a single model's test time and convenience.

\begin{figure*}[t]
  \centering
  \includegraphics[width=0.8\linewidth]{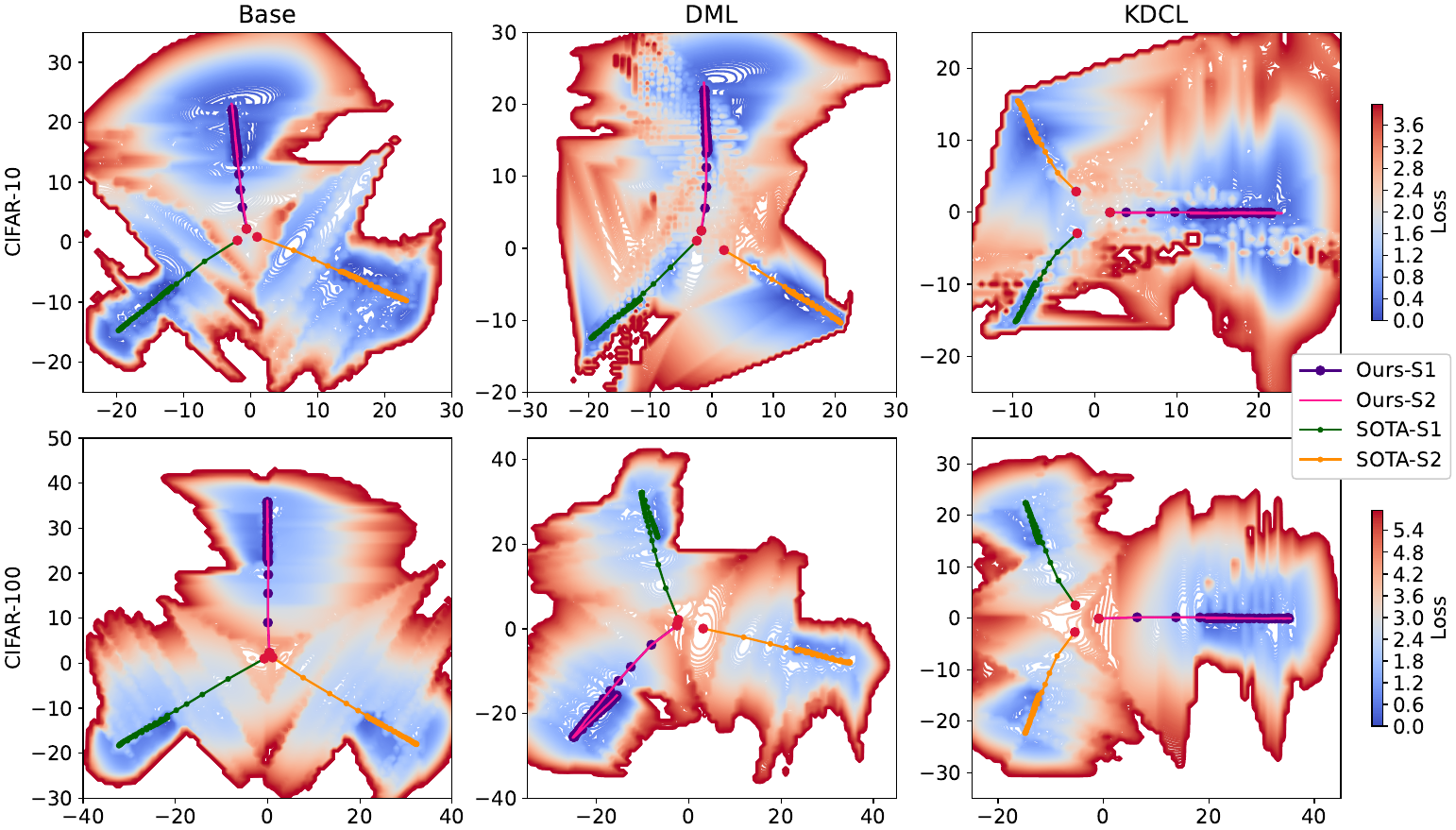}
  \caption{
  The loss landscape visualization of three methods~(Base, DML \cite{zhang2018deep}, and KDCL \cite{guo2020online} from left to right) compared with our method on two datasets~(CIFAR-10 and CIFAR-100 \cite{krizhevsky2009learning} from top to bottom).
  \textcolor{Plum}{Ours-S1} and \textcolor{VioletRed}{Ours-S2} are the two students obtained by our method, and \textcolor{ForestGreen}{SOTA-S1} and \textcolor{BurntOrange}{SOTA-S2} are the two students obtained by other methods, both of which are ResNet32 \cite{he2016deep} trained by the same settings.
  The x-axis and y-axis represent the values of model parameters by the PCA dimension reduction algorithm~\cite{mackiewicz1993principal}.
  Each sub-diagram shows four students who start from the initial point~(Red points in the center) and converge to three basins along different loss trajectories.
  }
  \label{fig:figure3}
\end{figure*}

\subsection{Generalization Measurement} \label{generalization:measure}
In this section, we conduct the generalization evaluation experiments using two traditional generalization measurements: the flatness of loss minima~\cite{keskar2016large,hochreiter1997flat} and the difference between the perturbed and empirical loss~\cite{neyshabur2017exploring,langford2001not}.

\subsubsection{Loss Landscape Visualization}
We visualize multiple students obtained by various methods onto a single diagram to intuitively compare their differences in the flatness of the loss landscape. 
Specifically, we construct a set of high-dimensional vectors obtained by flattening students' parameters and use the PCA dimension reduction algorithm \cite{mackiewicz1993principal} to generate the two-dimensional coordinates.
As shown in~\cref{fig:figure3}, the methods are Base~(two students are independently trained only by the CE loss), DML \cite{zhang2018deep}, and KDCL \cite{guo2020online} from left to right, and the color bar represents the training loss.
Here we only display the above three methods because other OKD baselines introduce extra modules or branches, resulting in the vast difference in parameter quantities and thus failing the landscape comparison by PCA.

As each sub-diagram in \cref{fig:figure3} shows, our students converge to a broader and flatter basin (thus superior generalization performance). 
In contrast, other students converge to two sharp basins, resulting in a difference in generalization and impairing the overall stability of performance.
Although students with different parameters are preferable for ensembles, which require diversified predictions to improve robustness, our approach differs from these ensemble methods.
Assuming that our students fall into different basins, direct parameter hybridization can easily break down due to the high nonlinearity of deep neural networks.
Therefore, our two students' loss trajectories are very close in the landscape, caused by the regular fusion operation~(\cref{eq:student-fusion}) shortening students' distance.
Considering our work aims to get one parameter that performs well in various scenarios, our students receiving different augmentations can become diverse quickly during training.
Our method pulls them in the same direction, acting as strong regularization and improving the generalization.
Overall, our OKDPH flattens loss minima and achieves the expected effect.

\begin{table*}[htbp]
\setlength{\tabcolsep}{3pt}
\begin{center}
\small
\begin{tabular}{l|l|ccccccc|ccccccc}
\toprule
\multirow{2}{*}{\textbf{Dataset} } & \multirow{2}{*}{\textbf{Setting} } &  \multicolumn{7}{c|}{\textbf{ResNet32}} & \multicolumn{7}{c}{\textbf{VGG16}} \\
&  & \footnotesize \textbf{DML} & \footnotesize \textbf{ONE} & \footnotesize \textbf{KDCL} & \footnotesize \textbf{OKDDip}  & \footnotesize \textbf{FFL} & \footnotesize \textbf{PCL} & \footnotesize \textbf{Ours}
& \footnotesize \textbf{DML} & \footnotesize \textbf{ONE} & \footnotesize \textbf{KDCL} & \footnotesize \textbf{OKDDip}  & \footnotesize \textbf{FFL} & \footnotesize \textbf{PCL} & \footnotesize \textbf{Ours} \\
\midrule
\multirow{3}{*}{\textbf{CIFAR-10}} 
& \textbf{Noisy} & 83.02 & 83.44 & 82.95 & 83.70 & 83.38 & 83.18 & \textbf{84.27} 
                & 83.96 & 83.09 & 83.91 & 83.34 & 83.00 & 83.07 & \textbf{84.93} \\
& \footnotesize \textbf{10\%} & 81.48 & 81.64 & 81.15 & 81.50 & 81.52 & 81.13 & \textbf{81.95} 
                                & 80.56 & 76.64 & 81.49 & 78.17 & 79.26 & 79.44 &  \textbf{83.40} \\
&  \textbf{1\%} & 38.66 & 41.08 & 41.20 & 38.68 & 38.55 & 33.17 & \textbf{44.14}
                & 41.03 & 35.93 & 42.79 & 35.75 & 40.28 & 39.73 & \textbf{43.54} \\
\midrule
\multirow{3}{*}{\textbf{CIFAR-100}} 
& \textbf{Noisy} & 51.84 & 52.07 & 50.06 & 52.16 & 50.49 & 50.43 & \textbf{52.89}
                & 51.59 & 49.81 & 49.38 & 49.28 & 48.62 & 50.61 &  \textbf{53.63} \\
& \footnotesize \textbf{10\%} & 39.99 & 40.92 & 39.05 & 40.88 & 39.20 & 41.54 & \textbf{41.83} 
                            & 38.65 & 28.21 & 38.19 & 29.09 & 32.86 & 28.99 & \textbf{42.59} \\
&  \textbf{1\%} & 8.70 & 8.13 & 9.40 & 7.64 & 8.80 & 9.42 & \textbf{9.71} 
                & 7.73 & 5.18 & 7.38 & 5.91 & 6.62 & 5.27 & \textbf{8.32} \\
\bottomrule
\end{tabular}
\caption{Top 1 accuracy~(\%) comparison of several OKD methods in the context of noisy data~(\textbf{Noisy}) and limited data~(Sampling \textbf{10\%} and \textbf{1\%} of training data).
}
\label{Noisy}
\end{center}
\end{table*}

\subsubsection{Stability Analysis}
Except for the flatness of the loss landscape, stability analysis of models is also the primary tool for measuring generalization~\cite{neyshabur2017exploring,langford2001not}.
In this subsection, we evaluate the performance of several OKD techniques in two contexts of noisy and limited data.

Specifically, the Gaussian random noise is added to the images in the training data, as shown below:
\begin{equation}
    X = X + \rm Gussian(\mu, \lambda),
\end{equation}
where $X$ is an image in the training data, and $\mu$ and $\lambda$ are the mean and variance of the noise.
Here we set $\mu$ and $\lambda$ to $0$ and $1$ and calculate the accuracy on the same original test data.
Besides, We randomly select $10\%$ and $1\%$ of the data from the training set for model training, and the test set remains unchanged. 
It is worth emphasizing that taking $1\%$ of the training data on CIFAR-100 means that models can only see five images of the same kind, which is very challenging.

The results in \cref{Noisy} show that our method is far superior to other methods in well-known architectures. 
When $10\%$ data are used to train VGG16, our OKDPH is much better than other OKD methods, $1.91\%$ and $3.94\%$ higher than the suboptimal method on CIFAR-10 and CIFAR-100, respectively.
Compared with training on the complete dataset, our method shows more robust performance in the scenario of limited data, mainly due to the regularization effect brought by our parameter hybridization.
The overparameterization caused by various modules in other OKD methods will introduce more uncertainties, manifested as the phenomenon of overfitting, which reduces the generalization ability, thus leading to failure.

\subsection{Parameter Sensitivity} \label{parameter_sensitivity}
As shown in \cref{omega}, we explore the impact of four different hyperparameter~($\omega, \beta, \gamma, \Delta$) values on the performance.
We train two students with ResNet32 on CIFAR-10 for experiments and analyze the impact of one hyperparameter when the other three hyperparameters are fixed.
The fusion proportion and interval $\gamma,\Delta$ directly constrain the distances between students since effective parameter hybridization requires a high similarity of multi-model parameters.
The sub-tables in~\cref{omega} show that the too-high or too-low similarity will lead to poor performance.
When $\gamma=0.0$, the performance is the worst, only $94.17\%$, which reflects the role of the fusion operation between the HWM and students.

\begin{table}[htbp]
\setlength{\tabcolsep}{4pt}
\begin{center}
\small

\begin{subtable}{1.0\linewidth}
    \centering
    \begin{tabular}{l|ccccc}
    \toprule
    \boldsymbol{$\omega$} & \textbf{0.2} &  \textbf{0.4} &  \textbf{0.6} &  \textbf{0.8}  &  \textbf{1.0} \\
    \midrule
    \textbf{Acc} & 94.49 & 94.53 &  94.68 & \bf 95.01  &  94.51\\
    \bottomrule
    \end{tabular}
    \caption{Influence of $\omega$ with $\beta=0.8, \gamma=0.5, \Delta=\text{1e}$.}
\end{subtable}

\hspace{2em}

\begin{subtable}{1.0\linewidth}
    \centering
    \begin{tabular}{l|ccccc}
    \toprule
    \boldsymbol{$\beta$} & \textbf{0.0} &  \textbf{0.2} &  \textbf{0.5} &  \textbf{0.8}  &  \textbf{1.0} \\
    \midrule
    \textbf{Acc} & 94.73 & 94.79 & 94.83 & \bf 95.01 & 94.59\\
    \bottomrule
    \end{tabular}
    \caption{Influence of $\beta$ with $\omega=0.8, \gamma=0.5, \Delta=\text{1e}$.}
\end{subtable}

\hspace{2em}

\begin{subtable}{1.0\linewidth}
    \centering
    \begin{tabular}{l|ccccc}
    \toprule
    \boldsymbol{$\gamma$} & \textbf{0.0} &  \textbf{0.2} &  \textbf{0.5} &  \textbf{0.8}  &  \textbf{1.0} \\
    \midrule
    \textbf{Acc} & 94.17 & 94.62 & \bf 95.01 & 94.74 & 94.56\\
    \bottomrule
    \end{tabular}
    \caption{Influence of $\gamma$ with $\omega=0.8, \beta=0.8, \Delta=\text{1e}$.}
\end{subtable}

\hspace{2em}

\begin{subtable}{1.0\linewidth}
    \centering
    \begin{tabular}{l|ccccc}
    \toprule
    \boldsymbol{$\Delta$} & \textbf{5b} &  \textbf{200b} &  \textbf{1e} &  \textbf{2e}  &  \textbf{5e} \\
    \midrule
    \textbf{Acc} & 94.73 & 94.81 & \bf 95.01 & 94.64 & 93.79\\
    \bottomrule
    \end{tabular}
    \caption{Influence of $\Delta$ with $\omega=0.8, \beta=0.8, \gamma=0.5$.}
\end{subtable}

\caption{Results~(\%) of OKDPH using ResNet32 with different values of four hyperparameters~($\omega, \beta, \gamma, \Delta$) on CIFAR-10, where b and e are abbreviations for batch and epoch, respectively.
}
\label{omega}

\end{center}
\end{table}

\subsection{Ablation Study} \label{ablation:analyse}
In this section, we analyze the contribution of different components in our OKDPH to the final performance, including the KD loss ${\mathcal{L}}_{kd}$, the fusion of HWM and students~(\cref{eq:student-fusion}), and the HWM's classification loss ${\mathcal{L}}_{ce}^{hwm}$.
\cref{Ablation} shows the accuracy and performance improvement in four settings of the models trained under the backbone of ResNet32.
As the basis of the whole distillation process, ${\mathcal{L}}_{kd}$ brings the greatest performance improvement.
It is necessary to fuse HWM's and students' parameters, which bring $0.28\%$ and $0.29\%$ improvements on CIFAR-10 and CIFAR-100, respectively.
Due to the contribution of the ${\mathcal{L}}_{ce}^{hwm}$, our method achieves greater than $95\%$ accuracy on CIFAR-10, breaking through the performance bottleneck and proving our idea's effectiveness.

Please refer to the supplementary materials for more experimental results, including experiments with training three or more students, extensive comparisons of generalization, and the display of hyperparameter values.

\begin{table}[t]
\setlength{\tabcolsep}{3pt}
\begin{center}
\small

\begin{tabular}{>{\centering\arraybackslash}p{0.1\textwidth}|cccc|>{\centering\arraybackslash}p{0.1\textwidth}}

\toprule
\textbf{Dataset} & \boldsymbol{${\mathcal{L}}_{ce}$} & \boldsymbol{${\mathcal{L}}_{kd}$}  & \textbf{Fuse} & \boldsymbol{${\mathcal{L}}_{ce}^{hwm}$} & \textbf{Acc~(\%)} \\
\midrule
\multirow{4}{*}{\textbf{CIFAR-10}} & \checkmark &  &  &  & 93.26 \\
 & \checkmark & \checkmark &  &  & 94.47~\textcolor{ForestGreen}{(+1.21)} \\
 & \checkmark  & \checkmark & \checkmark &  & 94.75~\textcolor{ForestGreen}{(+0.28)} \\
 &  \checkmark  & \checkmark & \checkmark & \checkmark & 95.01~\textcolor{ForestGreen}{(+0.26)} \\
\midrule
\multirow{4}{*}{\textbf{CIFAR-100}} & \checkmark &  &  &  & 72.76 \\
 &  \checkmark & \checkmark &  &  & 73.31~\textcolor{ForestGreen}{(+0.55)} \\
 &  \checkmark  & \checkmark & \checkmark &  & 73.60~\textcolor{ForestGreen}{(+0.29)} \\
 &  \checkmark  & \checkmark & \checkmark & \checkmark & 74.10~\textcolor{ForestGreen}{(+0.50)} \\
\bottomrule

\end{tabular}

\caption{Accuracy and performance improvement of four parts of our OKDPH using ResNet32 on CIFAR-10 and CIFAR-100.}
\label{Ablation}

\end{center}
\end{table}

\section{Conclusion}
In this paper, we aim to explicitly measure the generalization in OKD and propose OKDPH to promote flatter loss minima and more stable convergence of students. 
A sampled hybrid-weight model, \ie, HWM, is constructed in every training batch via the linear combination of all the students.
Then, we adopt the supervision loss of HWM to guide students to converge to a flatter loss minima.
Also, a novel fusion operation is designed to control the similarity of students to achieve effective parameter hybridization.
Experiments with various backbones and datasets prove that OKDPH performs significantly better than SOTA methods of two fields.
However, the limitation of OKDPH is that the parameter hybridization process can only be applied to homogeneous students.
In the future, we strive to eliminate the limitation of multiple models and extend our method as a general optimizer to achieve broader applicability.

\paragraph{Acknowledgements.}
This work is funded by the National Key Research and Development Project (Grant No: 2022YFB2703100), National Natural Science Foundation of China (61976186, U20B2066, 62106235, 62106220), Ningbo Natural Science Foundation (2021J189), Fundamental Research Funds for the Central Universities (2021FZZX001-23), Open Research Projects of Zhejiang Lab (NO. 2019KD0AD01/018), and Exploratory Research Project of Zhejiang Lab(2022PG0AN01).

{\small
\bibliographystyle{ieee_fullname}
\bibliography{egbib}
}

\end{document}